\documentclass[journal,twoside,web]{ieeecolor}
\usepackage{tmi}
\usepackage{amsmath,amsfonts}
\usepackage{algorithmic}
\usepackage{algorithm}
\usepackage{array}
\usepackage[caption=false,font=normalsize,labelfont=sf,textfont=sf]{subfig}
\usepackage{textcomp}
\usepackage{stfloats}
\usepackage{url}
\usepackage{verbatim}
\usepackage{graphicx}
\usepackage{cite}
\usepackage{amsmath,amssymb,amsfonts}
\usepackage{graphicx}
\usepackage{multirow}
\usepackage{threeparttable}
\usepackage{booktabs}
\usepackage{makecell}
\def\BibTeX{{\rm B\kern-.05em{\sc i\kern-.025em b}\kern-.08em
    T\kern-.1667em\lower.7ex\hbox{E}\kern-.125emX}}
\begin{document}

\title{Prefer-DAS: Learning from  Local Preferences and Sparse Prompts for  Domain Adaptive Segmentation of Electron Microscopy}

\author{Jiabao Chen, Shan Xiong, and Jialin Peng
\thanks{This work was partly supported by the NSFC under Grant 12471498. (Corresponding author: Jialin Peng) }
\thanks{J. Chen,  S. Xiong, and J. Peng are with the College of Computer Science and Technology, Huaqiao University, Xiamen 361021, China (e-mail: jialinpeng@hqu.edu.cn).}
}

\markboth{Journal of \LaTeX\ Class Files,~Vol.~14, No.~8, August~2021}%
{Shell \MakeLowercase{\textit{et al.}}: A Sample Article Using IEEEtran.cls for IEEE Journals}


\maketitle

\begin{abstract}
Domain adaptive segmentation (DAS) is a promising paradigm for efficiently delineating intracellular structures from various large-scale electron microscopy (EM) data without requiring extensive annotated data  in each  domain.  However,   current unsupervised domain adaptation (UDA) strategies often show limited performance and produce inaccurate and biased predictions,  which impede their practical applications. In this study, we present a more realistic yet annotation-efficient setting where we utilize sparse points and local  preferences as weak labels in the target domain. Specifically, we develop Prefer-DAS, which pioneers sparse promptable learning and  local preference alignment. The  Prefer-DAS is a promptable multitask model that integrates self-training and prompt-guided contrastive learning. Unlike SAM-like methods,  Prefer-DAS allows for the use of full, partial, and even no point prompts during both training and inference stages and thus enables both automatic and interactive segmentation. To refine spatially defective and biased segmentation, we introduce Local direct Preference Optimization (LPO), plug-and-play solution for aligning with spatially varying human feedback. To address the case of missing human feedback, we also introduce Unsupervised Preference Optimization (UPO), which leverages self-learned preferences. As a general framework,  the Prefer-DAS model can effectively perform both weakly-supervised and unsupervised DAS, depending on the availability of points and human preferences.  Comprehensive experiments across four challenging DAS tasks demonstrate that our model outperforms SAM-like methods as well as unsupervised and weakly-supervised DAS methods in both automatic and interactive segmentation modes, highlighting strong generalizability and flexibility. Additionally, the performance of our model is very close to or even exceeds that of supervised models. 
\end{abstract}


\begin{figure}[t]
\centering
\includegraphics[width=\linewidth]{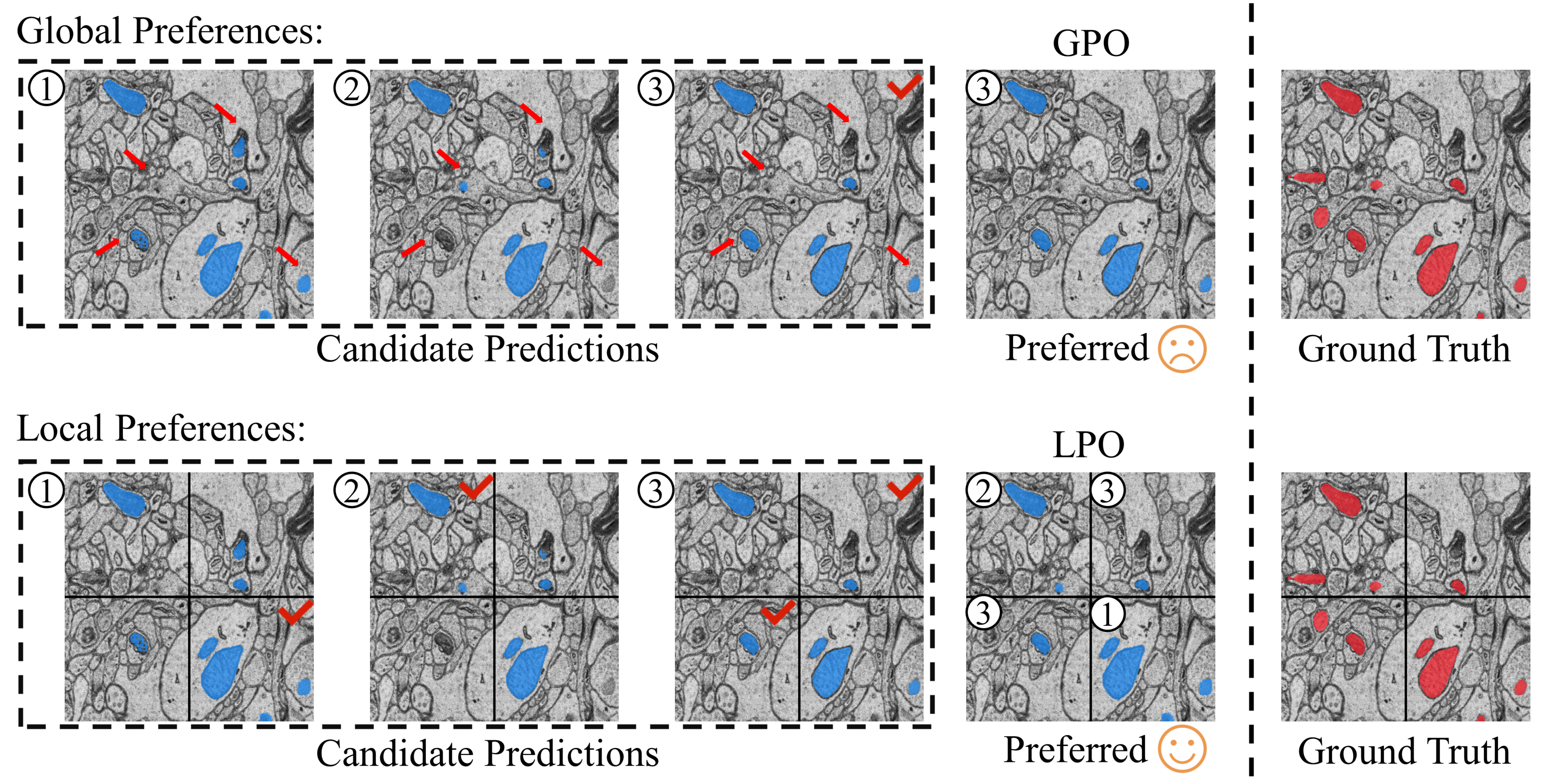}
\caption{A visualization of local human preferences. While selecting  preferred segmentation at the image level usually poses challenges for human raters and results in \textit{reward misspecification} in preference optimization, we gather local preferences at the patch level. Here, an image of small size and 2$\times$2 patches are used solely for illustrative purposes.  } \label{fig1}
\end{figure}

\section{Introduction}
\IEEEPARstart{R}{cently},  high-resolution volume electron microscopy (EM) images have become indispensable in biomedical research, experimental pathology, and diagnostic pathology for investigating intracellular structures and various diseases\cite{neikirk2023call}. A crucial initial step for analyzing intracellular structures across different cell types and states of cellular health using EM images is the precise and efficient segmentation of objects of interest, such as mitochondria, within diverse cellular contexts.  Recently, deep neural networks and large models \cite{kirillov2023segment,li2022dn} have revolutionized the field of automatic image segmentation and have been successfully applied in analyzing  EM images \cite{zhang2026masked,casser2018fast,peng2020unsupervised}.  
However, developing highly effective and efficient deep learning approaches for practical applications presents significant challenges. The amount of labeled data available and the granularity of annotation for training significantly influence model performance and generalization capabilities. Another major challenge comes from the sensitivity of the deep learning models  to data distribution variability. In the case of EM images,  manually annotating numerous instances of morphologically complex organelles requires substantial costs and expert knowledge.
Significant domain differences are typically observed across various EM images, which may come from diverse cell types of different organisms or be acquired using different microscopy techniques and imaging protocols.   Consequently, it is prohibitive to collect extensive  pixelwise expert annotations for each new domain.

Domain adaptive segmentation (DAS) focuses on adapting well-trained segmentation models from related domains (source domain) to the target domain, thereby presenting a promising strategy for reducing the necessity of annotating extensive datasets for each new domain. To alleviate  the issues associated with  the distribution shift or domain shift, various DAS approaches have been introduced. A prevalent setting is Unsupervised Domain Adaptation (UDA), which operates under the assumption that the source data is fully labeled while the target data remains unlabeled. However, in the absence of label information for the target data, UDA methods frequently encounter difficulties in achieving satisfactory performance, particularly in complex tasks. Although Semi-supervised Domain Adaptation (SDA) can achieve boosted performance with sufficient pixel-wise labels on the target domain, it demands substantial annotation effort.

To mitigate these challenges, we propose leveraging weak labels on the target domain and conducting Weakly-supervised Domain Adaptation (WDA), thereby significantly reducing annotation costs and the need for strong expert knowledge while achieving competitive performance. In this study, we consider three forms of coarse labels as weak supervision for model training:  1) Type I:  sparse center point annotation, which is particularly aligned with EM image segmentation or other cell image segmentation; 2) Type II:  patch-level human preferences, which are local human perceptual preferences  regarding
different segmentation predictions  and are suitable for both biomedical and natural image segmentation tasks; 3) Type III: self-learned preferences for correcting specific class of biased segmentation, such as misaligned boundaries.  Importantly, compared to pixelwise annotation and fully annotating all center points for each object instance in EM images, randomly annotating sparse points on a small number of object instances requires fewer expert resources and can usually be easily completed by non-experts.   

Recently, foundation segmentation models \cite{kirillov2023segment,ali2025review}, usually pre-trained on billion-scale datasets of natural images, have shown remarkably strong generalization abilities without training on specific targets. For instance, the Segment
Anything Model (SAM) \cite{kirillov2023segment},  designed to segment target objects  based on user prompts, has demonstrated impressive performance on various segmentation tasks. Particularly, the SAM model has intrigued a new shift towards developing more flexible, user-oriented segmentation paradigms, such as promptable and interactive models, aiming to enhance segmentation performance by incorporating human interaction during either the training stage or the inference stage.  The SAM treats human interactions, such as points, boxes, or masks, as prompts to guide segmentation during both the training and inference stages. The promptable segmentation models pave the way for the longstanding interactive segmentation, which can be responsive to user intention or progressively refine the segmentation. However, SAM relies on prompting for each instance to segment all object instances, making it particularly challenging to segment numerous organelle instances from EM images.  SAM still struggles with domain shifts and usually shows low performance on medical image tasks \cite{ali2025review}, especially with point prompts, due to the lack of medical knowledge, ambiguous boundaries,  and complex shapes. To enhance the performance, several studies \cite{zhang2024improving,cheng2023sam,wu2023medical}, have proposed modifying or fine-tuning  SAM using medical data, such as SAM-Med2D \cite{cheng2023sam}, and Med-SAM Adapter \cite{wu2023medical}.  Furthermore, SAM, like other supervised models, just learns from paired image-annotation datasets and does not incorporate human feedback during training.

\begin{figure}[t]
\centering
\includegraphics[width=0.49\textwidth]{Fig2.pdf}
\caption{Illustration of our Prefer-DAS model for weakly-supervised domain adaptive segmentation. Our framework comprises two stages. The first stage performs promptable domain-adaptive segmentation for UDA or WDA segmentation, depending on the availability of sparse points; the second stage performs locally supervised or unsupervised preference learning, depending on the availability of human preferences. } \label{fig2}
\end{figure}

Recently, the issue of misalignment between model outputs and human intention has attracted significant research attention \cite{christiano2017deep},  particularly concerning  Large Language Models (LLMs), which usually generate hallucinatory and biased responses.  The issue of misalignment also exists in DSA tasks,  largely due to the substantial domain gap and limited annotation on the target domain.
 For LLMs, post-training reinforcement learning (RL) strategies such as  Direct Preference Optimization (DPO) \cite{rafailov2023direct} have been widely adopted to learn from human preferences, which serve as reward signals for model fine-tuning. The DPO strategy refines LLMs directly by utilizing human preferences, without relying on explicit reward modeling. Recently, DPO has also been applied in visual tasks, such as integrating with diffusion models for generating physically plausible images \cite{wallace2024diffusion}. 
However, human preference alignment for image segmentation with DPO remains largely underexplored.  
Although preference optimization has been explored in \cite{konwer2025enhancing} for enhancing the SAM-based semi-supervised segmentation, directly using DPO with image-level preferences, which we refer to as global preference optimization (GPO), is suboptimal. Given the complexity of the segmentation task and segmentation predictions, a single scalar rating score typically fails to accurately represent the quality of a segmentation prediction.  This can lead to what is known as \textbf{reward misspecification} \cite{tien2022causal} in preference optimization.  Moreover, for a human rater, selecting the preferred segmentation from multiple candidates for large-scale cellular images—which contain many objects of interest and rich content— can be more cumbersome and error-prone. A simple example is shown in  Fig. \ref{fig1}, where all three candidate predictions are globally imperfect. In contrast, conducting local ratings is more reasonable and user-friendly for human annotators, requiring significantly less annotation effort. Thus, 
we introduce patch-level preferences and propose enhancing our initially trained segmentation model with novel \textbf{Local direct Preference Learning (LPO)}, which essentially utilizes multiple rating scores for each image (see Fig. \ref{fig1}).  

In this study, we introduce  Prefer-DAS, a preference-guided promptable transformer model designed for domain-adaptive segmentation under incomplete and coarse-grained annotations, as shown in Fig \ref{fig2}.   As a promptable segmentation model, our Prefer-DAS is flexible and can effectively utilize both full and partial prompts during training and testing, enabling it to conduct both automatic and interactive segmentation.  The proposed model employs pseudo-prompt learning and multitask learning to address  the challenge of label scarcity associated with sparse points. Additionally, it incorporates prompt-guided contrastive learning to enhance the learning of discriminative features.  Unlike SAM-like models \cite{kirillov2023segment,wu2023medical}, our model does not assume the availability of prompts for each object of interest during testing and can conduct inference with any number of point prompts in a single pass.

As a preference-guided model, the proposed Prefer-DAS model can draw upon human ratings of the relative quality of different segmentation predictions to guide our model training and optimize the behavior of our segmentation model. 
The proposed preference learning module is formed as a plug-and-play post-training that can be combined with UDA and WDA. We address reward misspecification in the segmentation task by introducing local preference optimization (LPO), and unsupervised preference optimization (UPO).
Comprehensive experiments conducted on challenging benchmarks demonstrate the effectiveness of the
proposed approach.

A preliminary conference version appears in \cite{chen2025prompt}, where we introduced the Prompt-DAS model. This improved version introduces substantial methodological advancements, including local preference learning and unsupervised preference learning, along with extensive validations on more challenging benchmarks. Additionally, a more comprehensive literature review and new illustrations are presented.

\section{Related Work}
\subsection{Domain Adaptive Segmentation}
Most machine methods experience degraded performance when there is a discrepancy between the distributions of training and testing data. This decline in performance is also evident in cutting-edge large-scale models, such as the foundation segmentation model SAM \cite{kirillov2023segment}. While SAM demonstrates strong zero-shot generalization ability on various natural images with human prompts, it suffers significant performance drops on biomedical datasets \cite{ali2025review}, including EM images. Although several medical variants of SAM, such as SAM-Med2D \cite{cheng2023sam},  Med-SAM Adapter \cite{wu2023medical}, have been introduced, they still suffer from the large domain gaps among medical data.

Most domain adaptation methods tackle domain adaptive segmentation by aligning the source and target distributions in various feature spaces \cite{Outputspace}, or through a combination of these approaches \cite{peng2020unsupervised}. Recently, self-training, which focuses on exploiting unlabeled target data by estimating pseudo labels, has become a popular gain to existing adaptation methods due to its simple concept. 
For mitochondria segmentation from EM images, the 2D model DAMT-Net \cite{peng2020unsupervised} jointly learns domain-invariant visual and geometrical features using adversarial learning and an autoencoder. The DA-ISC \cite{wu2021uncertainty} is a 2.5D method that integrates output-space adversarial learning  with a novel intersection consistency technique. The UALR method \cite{wu2021uncertainty} is a self-training-based method that rectifies noisy labels by estimating their uncertainty. Furthermore,  the CAFA \cite{yin2023class} method enhances self-training through class-aware feature alignment. Additionally, our previous work, WDA-Net \cite{qiu2024weakly}, also conducts WDA segmentation using novel sparse points as the weak supervision.  Despite the significant advances in UDA and WDA tasks, prior methods typically can not accept human feedback and interaction and suffer from imprecise and biased segmentation. Our current work introduces the first local preference-guided WDA model, which can also utilize sparse point prompts as the WDA-Net \cite{qiu2024weakly}.

\subsection{Learning with Visual Prompts}
Beyond automatic segmentation, a longstanding research direction is to develop models that can leverage user interaction, which is particularly beneficial for medical image analysis. The SAM model \cite{kirillov2023segment}, which was pretrained on 1 billion masks on 11  million natural images, conducts interactive segmentation by using user-provided  bounding boxes, points, and masks as inference prompts, showing remarkable performance across various zero-shot tasks. Typically, SAM's effectiveness is influenced by the type of prompts \cite{ali2025review} and is sensitive to the randomness associated with point and box prompts. Generally, the performance of SAM with point prompts is not as strong as with box prompts. A significant limitation of the SAM model is that it requires at least one prompt for each object instance, which can complicate the segmentation of images that contain a large number of objects, such as cellular images.    
In this study, we introduce a promptable model that allows for inference with full, partial, or no point prompts for all target predictions in a single pass.

\subsection{Learning from Human Preferences}
Various preference learning strategies have shown notable success when integrated with LLMs, large vision-language models (LVLMs), and diffusion models.  The POPEN model \cite{zhu2025popen} leveraged feature-similarity-based DPO to reduce hallucinations in LVLM-based reasoning segmentation with text input. Konwer \textit{et al.} \cite{konwer2025enhancing} improved SAM with DPO for semi-supervised segmentation of medical images, in which global ratings on the quality of candidate segmentation are used. While globally rating and ranking the quality of two or more segmentation predictions is challenging and often less meaningful, we explore local preference learning, which allows us to efficiently 
fine-tune our segmentation models using limited preference
data. To further reduce the annotation effort, we also introduce self-learned preferences as the proximity to human preferences, and develop an unsupervised DPO.

\section{Preliminaries of DPO}
\textbf{Single negative}. The DPO typically maximizes the generation probabilities of the pairwise preferences $\mathcal{D}_h=\{(x,y_p,y_d)\}$, consisting of one preferred prediction $y_p$ and one dispreferred prediction $y_d$ (single negative).  Based on the Bradley-Terry (BT) model\cite{bradley1952rank}, the 
preference model is
\begin{equation} 
P\left(y_p\succ y_d|x\right)=\frac{\exp(R\left(x,y_p\right))}{\exp(R\left(x,y_p\right))+\exp(R\left(x,y_d\right))} \label{eq.1}
\end{equation}
where $R(\cdot,\cdot)$ is the reward model that measures
how well a prediction $y$ meets human preference for  the given input $x$.
There is a connection between the
reward and the optimal RLHF policy $p^*_{\theta}(y|x)$: 
$
R\left(x,y\right)=\beta \log p^*_{\theta}(y|x)/p_{0}(y|x)+\beta \log Z(x) 
$,
where $Z(\cdot)$ is the unknown partition function, $p_{0}$ is the  reference  policy, and $\beta$ is a positive hyperparameter.
The standard DPO learns the optimal policy by minimizing the  negative log-likelihood loss of Eq. \ref{eq.1},

\textbf{Multiple negatives}.  When multiple negatives  $\{y_{d_j}\}_{j\in I_d}$  are paired with a preferred prediction $y_p$,  similar expressions of DPO  can be derived under the Plackett-Luce (PL) model \cite{chen2024softmax},
 \begin{equation}  
P\left(y_p\succ y_{d_j},\forall j \in I_d|x\right)=\frac{\exp(R\left(x,y_p\right))}{\sum_j \exp(R\left(x,y_{d_j}\right))}\label{eq.2}
\end{equation}

\section{Method}
\subsection{Overview}
\subsubsection{Problem Setting}
 This study investigates cross-domain segmentation under weak supervision. Specifically, we have a well-trained model on the source domain $\mathcal{D}^s=\{(x^s,y^s)\}$, which has full pixel-wise labels $y^s$, and a weakly-labeled target domain $\mathcal{D}^t=\{(x^t,w^t)\}$ with weak labels $w^t$. In addition to the \textit{sparse point labels}  $w^t=\bar{c}^t$ used in our preliminary version \cite{chen2025prompt}, we further introduce  human feedback on the training data, i.e., human preferences on different segmentation predictions,  as the weak supervision in the form of \textit{local preferences} $w^t=r^t$.  Rather than obtaining preferred segmentation by globally evaluating candidate segmentation predictions  $\{y_i^t\}_{i=1}^K$, as shown in Fig. \ref{fig1}, the local preference setting selects the preferred segmentation $y_p^t$ and dispreferred ones $\{y_d^t\}$ in a patch-wise way. 
In addition, we also devise \textit{simulated preferences} $w^t=\bar{r}_s^t$ as self-learned weak labels.

Our objective is to develop a high-performing model that is flexible enough to perform UDA and WDA in both automatic and interactive segmentation modes,  depending on the availability of point prompts and preferences. Additionally, the model should effectively align with spatially varying feedback.

\subsubsection{Model Overview}  Fig. \ref{fig2} illustrates the proposed Prefer-DAS, which encompasses an image encoder $f_E$, a point prompt encoder $f_P$  that processes $M\geq0$ points at once as inputs, a multitask decoder $f_D$ followed by a semantic segmentation head $f_S$, and a regression-based center-point detection head $f_R$. Additionally, the model includes a preference calibration module.  By default, our Prefer-DAS model utilizes 15\% sparse points as training prompts,  and  local human preferences for preference learning.
In scenarios where  NO human preferences  are available  during model training, our Prefer-DAS model reverts to our preliminary version, Prompt-DAS.  
Furthermore, when partial/full point prompts  are provided during the testing phase, our model  performs interactive  segmentation, denoted as Prefer-DAS+.

To address the issue of label scarcity on the target domain, we conduct pseudo-label learning for both the segmentation and detection tasks under the mean-teacher framework \cite{NIPS2017_MT}. The output of the detection head $f_R$ is used to provide prompts for the segmentation task. Furthermore, the segmentation head $f_S$ is guided by a prompt-based contrastive loss, enhancing the discriminability of prompt embeddings.

\subsection{Promptable Detection}
The proposed Prefer-DAS utilizes an auxiliary detection task to enhance the segmentation learning. 
The center point detection task is relatively easier than the dense segmentation task, particularly given sparse points as training prompts and partial supervision. While the joint learning of multiple tasks can implicitly boost the segmentation performance,   confident detection outputs are further employed to augment the ground-truth point prompts for the segmentation task. Following the teacher-student framework\cite{NIPS2017_MT}, pseudo-labels for the unlabeled regions are generated by selecting most highest local maxima points with a threshold from the predicted density map by the teacher model, which is updated by the exponential moving
average of the student network. Note that local maxima points can be identified through Non-Maxima Suppression. For the target data,  student network training is supervised by both the ground-truth sparse points and pseudo labels, and the $M$ sparse points are also used as training prompts. In the scenario of UDA, where there are no point annotations on the target domain, we use the estimated confident points from the prediction of the source model as the pseudo-sparse points.
For the source data, ground truth center points are used as the supervision, and randomly sampled $n_s$ center points are used as training prompts. 
\begin{equation}
\mathcal{L}_{\mathrm{det}}=\frac{1}{|\mathcal{D}^s|}\sum_{x^s}L_d(F_R(x^s),d^s)+\frac{1}{|\mathcal{D}^t|}\sum_{x^t}L_d(F_R(x^t),\hat{d}^t) \label{eq.3}
\end{equation}
where $F_R$=$f_R\circ f_D\circ f_E$, $L_d$ represents mean square error loss, and $\hat{d}^t$ represents the density map generated by the target pseudo labels and ground truth points. 

\subsection{Promptable Segmentation}
To alleviate label scarcity, we leverage pseudo-labeling in the teacher-student framework.  Thus, both source labels and target pseudo-labels are used to supervise the model training. 
\begin{equation}
\mathcal{L}_{\mathrm{seg}} = \frac{1}{|D^s|} \sum_{x^s}L_s(F_S(x^s), y^s) + \frac{1}{|D^t|} \sum_{x^t} L_s(F_S(x^t), \hat{y}^t) \label{eq.4}
\end{equation}
where $F_S$=$f_S\circ f_D\circ f_E$, $L_s$ represents the standard cross-entropy loss, $\hat{y}^t$ represents the pseudo labels generated by the teacher model on the target domain. 

Similar to the detection task, we also use points as prompts for segmentation. For the source domain, we use the $n_s$ points sampled for the detection task as the training prompts. Note that $n_s$ is a random number during training.  Since the target data only has a few points as the annotation, we propose to use both the estimated points from the detection output and ground-truth sparse points as training prompts to assist the segmentation training. The target point prompts are generated by selecting the highest local maxima points with a threshold from the predicted density map by the detection head.

\subsection{Prompt-guided Contrastive Learning (PCL)}
To learn more discriminative embeddings during pseudo-label learning, we further introduce contrastive learning with the guidance of prompts, which can provide representative features to distinguish mitochondria instances from the background organelle. As shown in the left figure of Fig. \ref{fig1}, our contrastive learning aims to pull the feature embeddings of the estimated foreground points closer to those of the ground-truth sparse points while simultaneously pushing away from the foreground embeddings from the background embeddings. An MLP layer $\phi$ is utilized before conducting contrastive learning. Let $z^t = \phi(f_D(f_P(p^t)))$ denote the embedding derived from the target domain point $p^t$. We employ an attention mask mechanism following DN-DETR~\cite{li2022dn} to prevent information leakage from PCL.

Queries are generated from pixels identified as foreground exhibiting a sufficiently high confidence. Utilizing the pseudo-labels produced by the teacher model, we select 3 points from each instance with a confidence greater than $\delta_{f}$, resulting in $N^q$ foreground prompt embeddings $\{z^t_i\}_{i=1}^{N^q}$. Concurrently, we identify $N^n$ points with a confidence level below $\delta_{b}$, resulting in $N^n$ background prompt embeddings $\{\mu_k^b\}_{k=1}^{N^n}$. Since mitochondrial instances display high similarity, we employ the average embedding of sparse point prompts as the sparse prompt embedding $\mu^f$. The prompt-guided contrastive loss is 
\begin{equation}
\mathcal{L}_{\mathrm{pcl}}=-\sum_{i=1}^{N^q} \log \left[ \frac{\exp\left(\mu^f \cdot z^t_{i} /\tau \right)}{\exp\left(\mu^f \cdot z^t_{i}/ \tau  \right) +\sum_{k = 1}^{N^n} \exp\left( \mu_{k}^b \cdot z^t_{i} / \tau \right)} \right]\label{eq.7}
\end{equation}
\subsection{Local Preference Learning}
Typically, the predictions
of a segmentation model is usually biased by the modeling techniques in use and the data itself. This issue is more severe with the domain gap 
 and only weak/sparse labels on the target domain.
 To address these issues with low annotation cost, we propose aligning the segmentation model with human ratings through preference learning. Our objective is to ensure that our segmentation model generates segmentation predictions preferred by human raters with a higher probability than segmentation predictions that are not preferred.
 More specifically, we fine-tune an initially trained model on human-labeled reference data with improved DPO strategies. Thus, our strategy is a plug-and-play method that is model-agnostic.  

\subsubsection{\textbf{L}ocal direct \textbf{P}reference \textbf{O}ptimization (\textbf{LPO})}  Given several candidate segmentation predictions, a significant challenge in constructing human preference data is selecting the best segmentation prediction. Images contain spatial information and often contain complex content.  Typically, a segmentation may perform better than other segmentations in certain local regions but worse in other local regions. Thus, it is usually prohibitive to directly select the preferred segmentation from candidate segmentation predictions of low quality. To address these issues,
we first introduce the LPO  for the segmentation task. Specifically, we split each input image $x$ and its predictions into $L\times L$ patches ($L$=3 in our experiments) and collect local preferences in a patch-wise way.
With multiple dispreferred predictions, the objective function of our  LPO under the PL model (Eq \ref{eq.2}) is as follows,
\begin{equation}
\begin{aligned}
\mathcal{L}_{\mathrm{LPO}}^{PL}=-\mathbb{E}_{\mathcal{D}_h}\bigg[\sum_{l=1}^{L^2}\log\sigma\bigg(-\log\sum_{j\in I_d}&\exp\bigg(\beta\log\frac{p_{\theta}(y_{p}^l| x^l)}{p_{0}(y_{p}^l|x^l)}\\-&\beta\log\frac{p_{\theta}(y_{d_j}^l|x^l)}{p_{0}(y_{d_j}^l|x^l)}\bigg)\bigg)\bigg] 
\end{aligned}\label{eq.8}
\end{equation}
where $\{y_{d_j^l}\}_{j\in I_d}$ are the dispreferred candidate perfections for the $l$ th image patch $x^l$ and $y_{p}^l$ is the preferred local prediction, and  $\beta$ controls the deviation from the reference policy.

\subsubsection{Candidate prediction generation} In this study, we pair one preferred segmentation with multiple negatives in our preference data. While there are diverse ways to generate candidate segmentation predictions, we use a simple yet effective threshold-based strategy. Given a probabilistic prediction $p(x^t)$ for an input image $x^t$, we generate $J$ binary segmentation predictions $\{y_j\}_{j=1}^J$ through a group of thresholds $\{\gamma_j\}_{j=1}^J$. The candidate predictions are further used to construct preference data in an image-level way  or a local patch-level way.
 
\subsection{Unsupervised  Preference Learning}
We further introduce \textbf{U}nsupervised direct \textbf{P}reference \textbf{O}ptimization (\textbf{UPO}) with self-learned supervision to address scenarios where no human preferences are available. For unsupervised and weakly supervised cross-domain segmentation, a typical segmentation defect in medical image segmentation is the imprecisely aligned boundary with the true object boundary, which may stem from the bias of the source domain annotations, domain gap, and ambiguity in object boundaries. 

To address the issues of imprecise segmentation boundaries, we introduce a self-learned pseudo label for the target domain. Specifically, given a coarse segmentation $\hat{y}^t_0$ for a target training image $x^t$, we refine it with an edge-based active contour model, which evolves contours (e.g., the initial segmentation boundaries) to track the boundaries of desired objects under the guidance of an edge indicator $
g\triangleq\frac{1}{1+|\nabla G_\sigma*x^t|^2}$
where $G_\sigma$ is a gaussian kernel, $\sigma$  determines the width of the gaussian kernel, $*$ is the convolution operation. For 
simplicity, we use a well-known level-set-based formulation, namely distance regularized level set evolution (DRLSE) \cite{li2010distance}, for the active contour model. Other more powerful methods in variational formulation can also be used.

By applying the active contour model on a coarse segmentation $\hat{y}^t_0$, we can obtain a refined segmentation $\Tilde{y}^t$, which can be used to simulate a human rater through ranking the Dice scores between $\Tilde{y}^t$ and other candidate segmentation predictions. Thus, the preferred prediction is selected as follows,
\begin{equation}
y_p^t= \arg\max_{y\in\{y_j^t\}} Dice(\Tilde{y}^t,y)
\end{equation}

\subsection{Overall Optimization}
 To train our model, we incorporate the  detection loss $\mathcal{L}_{\mathrm{det}}$, segmentation loss $\mathcal{L}_{\mathrm{seg}}$, contrastive learning loss $\mathcal{L}_{\mathrm{pcl}}$, as well as the preference learning loss $\mathcal{L}_{\mathrm{PO}}^{PL}$ with $\mathrm{PO} \in \{ {\mathrm{GPO}},
{\mathrm{LPO}},  {\mathrm{UPO}}
 \}$, depending on the availability of human preference data. The \textbf{GPO}  refers to direct preference learning with global preference data, and the UPO refers to unsupervised preference learning with the self-learned preference data.  The loss function for our Prefer-DAS model is as follows,
\begin{equation}
\mathcal{L}_{\mathrm{Prefer-DAS}}=\mathcal{L}_{ \mathrm{seg}}+\lambda_1\mathcal{L}_{\mathrm{det}}+\lambda_2\mathcal{L}_{\mathrm{pcl}}+\lambda_3\mathcal{L}_{\mathrm{PO}}^{PL}
\label{eq.11}
\end{equation} 
where $\lambda_1$, $\lambda_2$, and $\lambda_3$ are  tradeoff parameters. When $\lambda_3$=0, our  Prefer-DAS uses no preference learning and degenerates to our preliminary conference version, i.e., Prompt-DAS. By default, our Prefer-DAS uses 15\% points as training prompts. During the inference stage, only the segmentation head is kept.

Significantly, our model is a flexible framework and is capable of performing both WDA and UDA. Specifically, the Prefer-DAS model can conduct WDA when using sparse points,  human preferences, or both of them are available during model training. It can also conduct UDA without using sparse points and  human preferences, or with self-learned human preferences during model training.
Moreover, the Prefer-DAS model can conduct interactive segmentation when full/partial points are available during the inference stage.

\section{Experiments}
\subsection{Benchmark and Metrics}
\textbf{Lucchi++}. This dataset \cite{casser2018fast} is a re-annotation of the EPFL Hippocampus dataset \cite{lucchi2011supervoxel} by three experts. This  dataset was taken from the  CA hippocampus region of a mouse using Focused Ion Beam Scanning Electron Microscope (FIB-SEM) at a resolution of 5$\times$5$\times$5 $n m^3$. This dataset has two subsets of size 768$\times$1024$\times$165 for training and testing, respectively.

\textbf{MitoEM-Rat (R)}. This dataset is a subset of MitoEM \cite{wei2020mitoem}, acquired using multi-beam scanning electron microscopy (MB-SEM) at a resolution of 8$\times$8$\times$30 $n m^3$. The images in this dataset represent tissue taken from Layer II/III in the primary visual cortex of an adult rat.
  The dataset contains a training subset of size 4096$\times$4096$\times$400 and a testing subset of size 4096$\times$4096$\times$100,  featuring many large instances with complex morphologies and approximately 14.4k instances of mitochondria, significantly more than the Lucchi++ dataset.

\textbf{MitoEM-Human (H)}. This dataset is a subset of MitoEM \cite{wei2020mitoem} and was taken from the Layer II in the temporal lobe of an adult human
using Multi-Beam Scanning Electron Microscopy (MB-SEM) at a resolution of 8$\times$8$\times$30 $n m^3$.  The dataset contains a training subset of size 4096$\times$4096$\times$400 and a testing subset of size 4096$\times$4096$\times$100.  Compared to  \textbf{R}, \textbf{H} contains a significantly higher number of instances of mitochondria, particularly small mitochondria.  This variety in terms of shape and density makes this dataset a more challenging benchmark.

\textbf{ME2-Stem (Stem)}. The ME2-Stem data were provided by Jiang et al. \cite{jiang2025efficient}, and now this dataset is a subset of the MitoEM 2.0 Dataset.  ME2-Stem data were taken from the brain stem of a mouse using Serial Block-Face Scanning Electron Microscopy (SBF-SEM) at a resolution of 8$\times$8$\times$30 $n m^3$. ME2-Stem data consist of mixed neural and glial populations.
This dataset contains three 1000$\times$1000$\times$100-voxel volumes, designated for training, validation, and testing, respectively.

\begin{table*}[t]
\caption{Quantitative comparison. Leading UDA, WDA, and SAM-like approaches in both automatic and interactive modes are compared. Our Prefer-DAS is flexible to conduct both UDA and WDA, depending on the availability of the training point prompts and human preferences, and can utilize sparse point prompts during inference.  For interactive segmentation, indicated by a "+",  full center points for all mitochondria are used as testing prompts to fulfill the requirements of SAM-like methods.   } 
\centering
 \label{tab:1}
\begin{threeparttable}
\setlength{\tabcolsep}{1.4mm}
\begin{tabular}{l|l|c|cccccccccccccccc}
\toprule
\multirow{2}{*}{Types} &\multirow{2}{*}{Methods} & Training  &   &\multicolumn{3}{c}{H $\rightarrow$ R} & ~ 
&\multicolumn{3}{c}{R $\rightarrow$ H} & ~ &\multicolumn{3}{c}{H $\rightarrow$ Lucchi++} & ~ &\multicolumn{3}{c}{H $\rightarrow$ Stem}\\
\cmidrule{5-7}\cmidrule{9-11}\cmidrule{13-15}\cmidrule{17-19}
~ & &Point Prompts  & ~ &Dice  & AJI   & PQ &~ & Dice  & AJI   &PQ &~ & Dice  & AJI  &PQ &~ & Dice  & AJI  &PQ\\
\midrule
\multicolumn{14}{l}{\textbf{Automatic Segmentation Mode}}&\\
\midrule
\multirow{3}{*}{SAM-like} & SAM \cite{kirillov2023segment} 
& \multirow{4}{*}{0\%}  & 
& 32.0 & 14.3 & 30.0 &  
&  20.8 & 11.4 & 18.7 &  
& 21.5  & 10.3 & 11.3 &
& 19.0 & 0.9 & 0.3\\
 &SAM-Med2D \cite{cheng2023sam}  &   & 
& 15.9 & - & - & ~ 
& 22.5 &  - & - &  
&  23.1 & 12.7 & 2.6 &
& 27.9 & 13.0 & 3.3\\
  & Med-SAM Adapter\cite{wu2023medical}\tnote{\dag}&  ~ & ~ 
& 75.5 & 56.6 & 27.5 & ~ 
& 72.4 & 55.0 & 33.4 &  
&  64.6 & 44.8 & 19.5 &
& 27.8 & 12.9 & 7.1\\
\midrule
\multirow{7}{*}{UDA}  &UALR \cite{wu2021uncertainty}  & \multirow{7}{*}{0\%} & 
& 86.3& 71.6&53.7 & ~
& 83.8& 69.7&60.0  &  
&  81.3 & 68.1 & 63.2 &
&65.5 & 38.7 & 28.7\\
~ & CAFA \cite{yin2023class} & ~ &
& 89.5& 77.6& 68.9 & ~ 
& 85.4& 72.9& 59.4 &  
& 86.7 & 76.7 & 70.1 &
& 79.6 & 47.4 & 48.9\\
 ~  &DA-ISC \cite{huang2022domain}  & ~ & 
& 88.6& 75.7&65.8 & ~
& 85.6& 72.7&63.8 &  
& 85.2& 72.8 & 64.1 &
& 72.0 & 38.8 & 35.6\\
 &DAMT-Net\cite{peng2020unsupervised}  &   &  ~
& 88.7& 76.3&61.8 & ~
& 85.4& 72.3& 63.7 &  
& 86.6& 75.5 & 73.0 &
& 78.6& 48.0 & 53.5\\
& WDA-Net (UDA) \cite{qiu2024weakly}  & &
& 88.2& 74.5&59.0 & ~
&85.5 &72.3 &60.6 & 
& 85.6 & 74.3 & 71.9 &
& 74.5 & 48.2 & 44.0\\
&Prompt-DAS (UDA) \cite{chen2025prompt}   &  &
&92.4 &82.2 &74.3  & 
& 88.0& 76.6&68.1 &  
& 87.3 & 76.9 & 73.1 &
& 82.9 & 51.6 & 53.6\\
 & \textbf{Prefer-DAS (UPO) }  & &
& 93.1 & 83.5 & 76.1 & 
& 89.1 & 78.1& 70.1 &  
& 92.1 & 84.7 & 79.6 &
& 83.3 & 53.9 & 54.3\\
\midrule
\multirow{5}{*}{WDA} & WDA-Net \cite{qiu2024weakly} & \multirow{5}{*}{15\%} & 
& 91.7& 80.7&74.0 & ~ 
& 88.7& 77.6&67.8 &  
& 86.3 & 75.6 & 70.8 &
& 80.7 & 48.8 & 50.0\\
 &Prompt-DAS \cite{chen2025prompt} &   &
&  93.3  &  83.6 & 74.5 & ~ 
& 89.2 & 78.6 & 69.1 &  
&  88.2 & 78.3 & 74.8 &
& 82.9 & 51.5 & 53.6\\
&Prefer-DAS (UPO) &     &
& 93.7 & 84.4 & 77.5 & ~ 
& 90.1 & 80.1 & 71.2 &  
& 92.6 & 85.6 & 80.9 &
& 83.4 & 51.9 & 53.8\\
 &\textbf{Prefer-DAS (LPO) } &    & 
& 94.6 & 85.9 & 79.6 & ~ 
& 92.2 & 83.6 & 75.8 & ~ 
& 94.1 & 88.3 & 82.3 &
& 85.6 & 67.9 & 62.0\\
\midrule
\multicolumn{14}{l}{\textbf{Interactive Segmentation Mode}}&\\
\midrule
\multirow{3}{*}{\shortstack{SAM-like}}  & SAM\texttt{+}  &  \multirow{4}{*}{0\%}    & 
& 40.6 & 1.2 & 26.2 & ~ 
& 40.3 & 4.6 &  26.6 &  
&  81.1 & 68.1 & 67.6 &
& 39.0 & 12.6 & 15.4\\
  &SAM-Med2D\texttt{+} &  & 
& 72.6 & 55.6 &  39.7 & ~ 
&  78.1 &  61.2 & 42.2 &
&  78.8 & 63.4 & 35.9 &
& 68.7 & 42.9 & 22.4\\
 & Med-SAM Adapter\texttt{+}\tnote{\dag} &  & 
& 86.2 & 70.2 & 59.9 & ~ 
& 83.8 & 68.1 & 59.0 &  
& 86.8 & 74.0 & 56.8 &
& 49.7 & 9.6 & 18.2\\
\midrule
 \multirow{5}{*}{\shortstack{WDA}}  & WeSAM\texttt{+} \cite{zhang2024improving}\tnote{$\ddag$} & \multirow{5}{*}{15\%}   & 
& 89.9 & 79.6 & 73.9 & ~ 
& 82.3 & 66.0 & 65.5 &  
& 83.2 & 69.0 & 63.0 &
& 44.9 & 7.3 & 17.7\\
 &Prompt-DAS\texttt{+} &   & 
&  93.5  & 84.4 & 74.2 & ~ 
& 90.8 & 81.5 & 72.3 &  
& 88.7 & 79.1 & 74.8 &
& 83.8 & 52.0 & 53.6\\
  &{Prefer-DAS (UPO)}\texttt{+} &   & 
& 94.3 & 85.4 & 79.2 & ~ 
& 90.9 & 81.2 & 71.6 &  
& 93.4 & 87.1 & 82.5 &
& 84.2 & 52.5 & 53.9\\
  &\textbf{Prefer-DAS (LPO)}\texttt{+} &   & 
& 95.1 & 86.7 & 81.6 & ~ 
& 92.8 & 84.6 & 77.4 & ~ 
& 94.7 & 89.2 & 84.4 &
& 86.3 & 68.6 & 62.0\\
\midrule
\multicolumn{14}{l}{\textbf{Segmentation Upper Bound}}&\\
\midrule
Oracle & Fully-Supervised Model &  \multirow{1}{*}{-} & 
& 94.6& 86.4&  79.2 & ~ 
& 92.6& 84.6& 75.8 &  
& 95.1 & 90.1 & 85.8 &
& 84.8 & 71.9 & 71.9\\
\bottomrule
\end{tabular}
\begin{tablenotes}[flushleft] 
       \item \tnote{\dag} Fine-tuning using the source data
        \item \tnote{$\ddag$} Fine-tuning using the source data and  target data with 15\% sparse point labels
     \end{tablenotes} 
\end{threeparttable} 
\end{table*}

\textbf{Evaluation metrics}. Following \cite{qiu2024weakly,chen2025prompt}, we utilize the semantic-level Dice and the instance-level Aggregated Jaccard Index (AJI), as well as Panoptic Quality (PQ).

\subsection{Implementation Details}
All the experiments are conducted
 on a single NVIDIA GeForce RTX 4090 GPU. We use DINO~\cite{Caron_2021_ICCV} as the backbone of our image encoder $f_E$ and use the pretrained ViT-S/8 \cite{Caron_2021_ICCV} to initialize the model parameters. Our decoder $f_D$ follows a design similar to SAM, with the addition of masked self-attention and masked cross-attention mechanisms to prevent information leakage. In contrast to SAM, our prompt encoder employs standard positional embeddings. The MLP module is identical to that used in SAM and other standard Transformer architectures. In the training stage 1 (Prompt-DAS stage), the model is trained for 16k iterations with a batch size of 2 and an initial learning rate of $1\times10^{-5}$. For a fair comparison, the same data augmentations as those in WDA-Net \cite{qiu2024weakly} were used. Randomly cropped image
patches of  $384\times384$ were used for training. In the contrastive loss,  we set $N^n$=256, $\delta_{f}$=0.9, $\delta_{b}$=0.1, and $\tau$=0.5. In the first stage,  we set $\lambda_1$=1e-3, $\lambda_2$=1e-3, and $\lambda_3$=0 in the overall loss.
  In the training stage 2 (Preference Learning stage), the model is initialized from the student model obtained in the previous stage and fine-tuned for 16k iterations with a batch of 4 and an initial learning rate of $5\times10^{-6}$. In both stages, we optimize the model using the AdamW optimizer with a polynomial learning rate decay of power 0.9.  We applied a group of 5 threshold values for candidate prediction generation: $\{0.3, 0.4, 0.5, 0.6, 0.7\}$. 
 For the preference optimization loss, we use $\beta$=1 and $\sigma$=1.
 In this stage,  we set $\lambda_1$=0, $\lambda_2$=0, and $\lambda_3$=1 in the overall loss.

\subsection{Quantitative Comparison} 
We compare our model with leading methods in both automatic and interactive segmentation modes. As shown in Table \ref{tab:1}, four settings of our models with different preference learning strategies are compared: Prompt-DAS, Prefer-DAS (UPO), and Prefer-DAS (LPO). The Prompt-DAS model is essentially  Prefer-DAS without preference learning and was introduced in our preliminary conference paper \cite{chen2025prompt}. Additionally, the supervised model is trained on pixel-wise labeled target data.

\textbf{Comparison with general/specialized foundation models}.  We begin by comparing our model with SAM \cite{kirillov2023segment}, which is capable of operating in both automatic and interactive modes with instance-wise point or box prompts. As shown in Table\ref{tab:1},  SAM  exhibits very low performance without model adaptation, primarily due to the significant differences between natural images and the EM images. While the SAM+ model, which utilizes a point prompt for each instance,  shows improved results over SAM without interactions,  its performance remains very low because of the shifted data distribution and the ambiguous boundaries of mitochondria in EM images. 
As shown in many studies \cite{ali2025review}, the SAM/SAM+ relies on sharp edges to delineate objects. 
Similar performance results are observed with the SAM-Med2D \cite{cheng2023sam}, which was trained by finetuning SAM with 4.6M medical images. When using full-point prompts, SAM-Med2D+ achieves substantial performance improvements—over 40\% in Dice for all three tasks. 

By finetuning on  the labeled source domain (indicated with $\dag$), the Medical SAM Adapter$^\dag$ \cite{wu2023medical}  and its interactive version demonstrate improved results across all four cross-domain segmentation tasks. Among the compared SAM-like methods, WeSAM, using both the labeled source data and 15\% sparse points on target for adaptation, obtain the best performance. 
However, these  models still exhibit a large performance gap compared to the supervised model and our Prefer-DAS.

Additionally, the requirement of point prompts on all instances during inference by these SAM-like models is a significant limitation for practical use.  In contrast, our model achieves strong performance with either no point prompts or just a few. Furthermore, our Prefer-DAS (LPO), without using any testing interaction, outperforms the supervised model by +0.8\% in Dice on the Human$\rightarrow$Stem task and shows only minor performance gaps of +0.0\%, -0.4\%, and -1.0\% in Dice on the other three domain-adaptation tasks, respectively, demonstrating its strong performance. With full point inference prompts, our Prefer-DAS+ model surpasses the supervised model in three of the four tasks,  primarily due to the effective joint usage of the source and target data.

\begin{figure*}[t]
\centering
\includegraphics[width=0.95\textwidth]{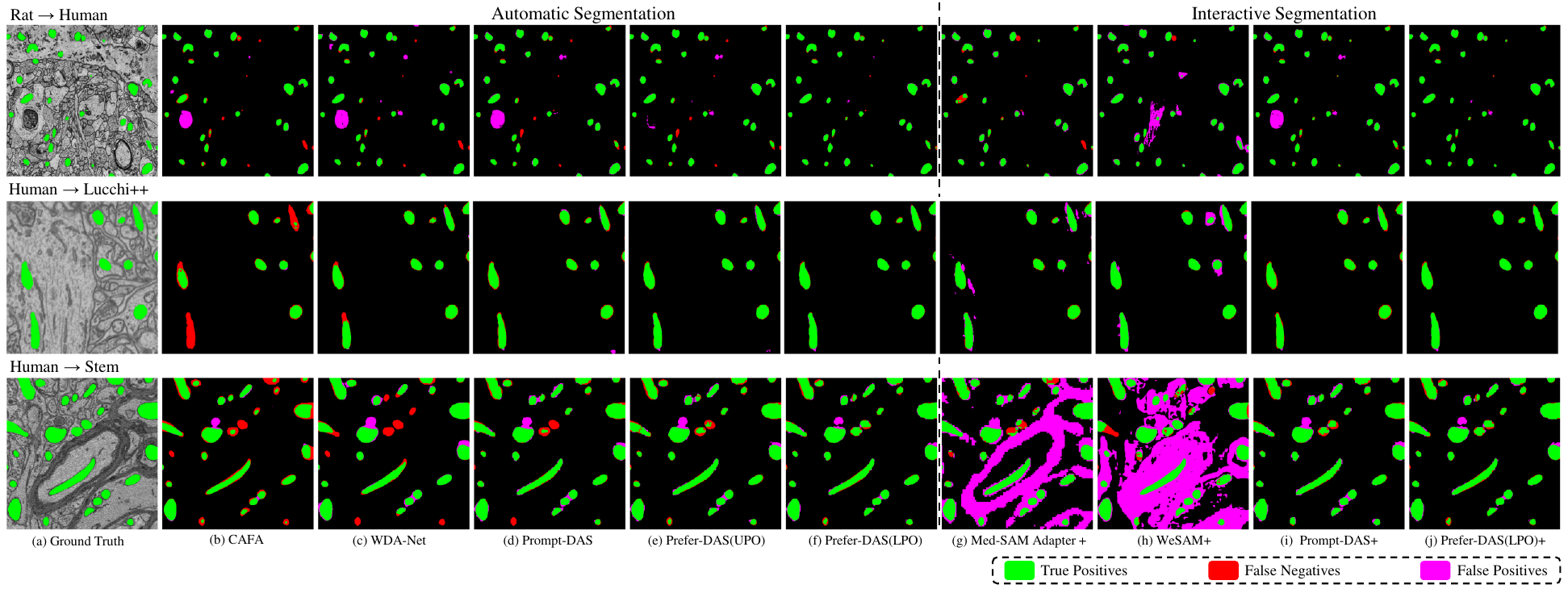}
\caption{Visual comparison of segmentation results in both automatic and interactive modes. Despite the Med-SAM Adapter+ and WeSAM+ being fine-tuned with the source EM data and utilizing full point prompts during inference,  they still show severe false positive segmentation. } \label{fig3}
\end{figure*}

\textbf{Comparison with UDA models}.  When using no labels on target data, our Prompt-DAS model performs UDA segmentation and is referred to as  \textbf{Prompt-DAS (UDA)}. Notably,  our \textbf{Prefer-DAS (UPO)}, which employs only self-learned preferences during training, also conducts UDA segmentation. As shown in Table \ref{tab:1}, our model in the UDA mode, i.e., Prompt-DAS (UDA), outperforms all other UDA methods, including various self-training-based methods, WDA-Net (UDA) \cite{qiu2024weakly}, UALR \cite{wu2021uncertainty},  and CAFA \cite{yin2023class}. Integrating UPO to Prompt-DAS (UDA), our Prefer-DAS (UPO) model obtains a performance improvement of 0.7\%, 1.1\%, 4.8\%, and 0.4\% in Dice scores for the four respective tasks. Notably,  larger performance gains are observed in terms of the AJI measure, indicating a better ability to correctly segment more mitochondria.

\textbf{Comparison with WDA models}. Firstly, as shown in Table \ref{tab:1},   using additional 15\% point prompts as weak labels for training results in significant performance improvements for the WDA-Net, Prompt-DAS, and  Prefer-DAS (UPO) models, particularly regarding instance-level AJI and PQ measures. Secondly, our Prefer-DAS (UPO), which employs self-learned preferences, outperforms all  UDA methods, as well as the WDA methods that utilize the same point labels. Specifically,  our Prefer-DAS (UPO) surpasses  WDA-Net (15\%) by substantial margins of 3.7\%, 2.5\%, 10.0\%, and 3.1\% in AJI for the four tasks, respectively. Thirdly,  using the proposed local preferences, Prefer-DAS (LPO) obtains the best performance, showing significant performance improvements of 5.2\%, 6.0\%, 12.7\%, and 19.1\% in AJI over the WDA-Net. 
Our Prefer-DAS (LPO) also outperforms the proposed Prefer-DAS (UPO) by  1.5\%, 2.5\%, 2.7\%, and 16.0\% in AJI. 
These results highlight the effectiveness of the proposed unsupervised preference learning and local preference learning methodologies.  

In the interactive mode, our models can further obtain substantial performance improvements,  and our Prefer-DAS (LPO)+ takes the best performance among all methods in comparison on all four tasks. Notably,  our Prefer-DAS (LPO)+ even outperforms the supervised model on three of the four tasks,  demonstrating the effectiveness of our method.

\textbf{Visual comparison}. Fig. \ref{fig3} presents a visual comparison of our methods against competing methods in both automatic and interactive modes. Overall, the results highlight the superior performance of our Prefer-DAS with significantly reduced false positives and false negatives. In contrast, the  CAFA and our  Prompt-DAS exhibit a notable number of false negatives for the H$\rightarrow$R task and display more false positives for the R$\rightarrow$H task. With unsupervised preferences,  Prefer-DAS (UPO) can effectively reduce false positives and negatives in large areas. Meanwhile,   Prefer-DAS (LPO) shows the best visual performance, showing minimal false positives and false negatives.  Although using testing prompts can enhance segmentation, the visual improvement is marginal, particularly when the initial segmentation quality is already high. 

\begin{table}[t]
\caption{Ablation study of the proposed model.}
\centering
\label{tab:2}
\setlength{\tabcolsep}{1.2pt}
\begin{tabular}{lcccccccccccc}
\toprule
 &\multicolumn{2}{c}{Pseudo-labeling}  & Training  & \multirow{2}{*}{PCL} &\multirow{2}{*}{LPO}&&\multicolumn{2}{c}{H $\rightarrow$ R} & &\multicolumn{2}{c}{R $\rightarrow$ H}\\
   \cmidrule{2-3}  \cmidrule{8-9} \cmidrule{11-12}
   & Det.& Seg. & Prompts &  & && Dice & PQ  & & Dice & PQ \\
\midrule
I&~  & ~         & ~          & & ~ & ~ &  88.6 & 68.7 & ~ & 78.1& 55.5\\
II&\checkmark & ~          & ~         & && & ~   89.2 & 70.4 & ~ & 87.4 & 68.4 \\
III& ~          &\checkmark         &  ~ & ~ && & 89.5 & 70.0 & ~ & 87.7& 68.3\\
IV&\checkmark &\checkmark          & ~& ~ & ~ && 90.4 & 71.8 &~ & 88.5 & 68.8 \\
V&\checkmark &\checkmark   & \checkmark & & ~ & ~ & 92.7 & 74.1 &~ & 88.9& 69.0\\
Prompt-DAS&\checkmark &\checkmark   & \checkmark & \checkmark & & ~ & 93.3 & 74.5 &~ & 89.2& 69.1\\
\textbf{Prefer-DAS}&\checkmark &\checkmark   & \checkmark & \checkmark &   \checkmark&  & \textbf{94.6} & \textbf{79.6} &~ & \textbf{92.2 }& \textbf{75.8}\\
\bottomrule
\end{tabular}
\end{table}
\subsection{Ablation Studies}
Table \ref{tab:2} summarizes the results of ablation analysis of our model across two domain adaptation tasks. 
Specifically, we evaluate the contributions of our key components: 1) Detection Pseudo-labeling; 2) Segmentation Pseudo-labeling; 3) Using 15\% sparse points as Training Prompts; 4) PCL: prompt-based contrastive learning; and 5) LPO: the proposed local preference optimization that uses patch-level preference data.

As shown in  Table \ref{tab:2}, the baseline Model I, i.e., our source model, can be improved by incorporating pseudo-labeling for detection or segmentation tasks. Through conducting multitask learning,  Model IV outperforms Model I by 1.8\% and 
10.4\% in Dice scores for the two tasks, respectively. By using sparse points as training prompts, Model V achieves additional improvements of 2.3\% and 0.4\%  in Dice scores over Model IV for the two tasks, respectively. Furthermore,  the integration of PCL into Model V yields Prompt-DAS, which leads to an average performance increase of 0.45\% in Dice for the two tasks. By further integrating the LPO, our  Prefer-DAS obtains an improvement of 1.5\% and 3.0\% in Dice, and a remarkable improvement of 5.1\% and 6.7\% in PQ for the two tasks, respectively. These results confirm the effectiveness of the proposed components, especially the LPO.

 \begin{figure}[t]
\centering
\includegraphics[width=0.75\linewidth]{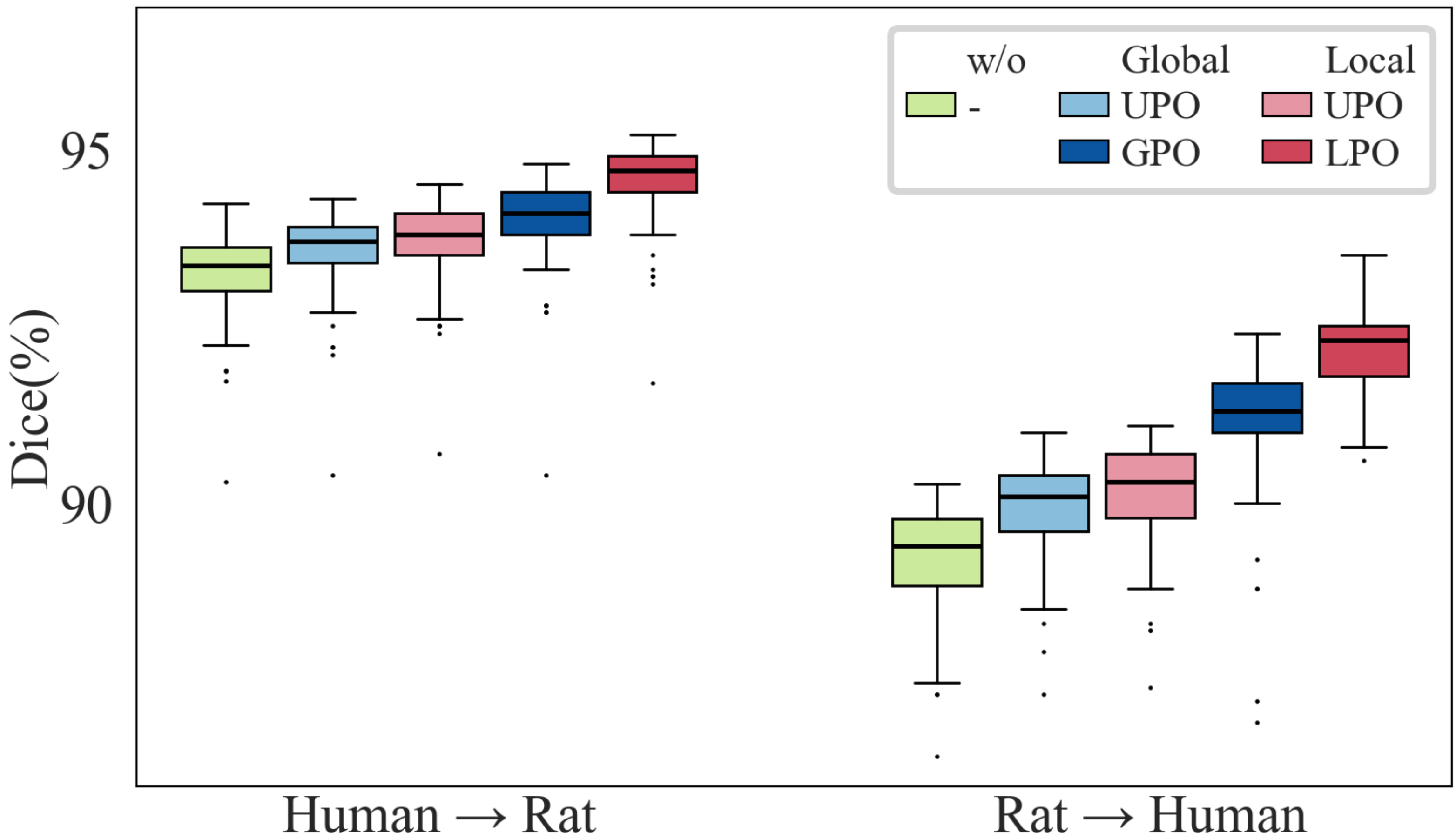}
\caption{Box plot of our Prefer-DAS using different DPOs. Global and local refer to image-level and patch-level preferences, respectively. "w/o" indicates the Prefer-DAS  using  no preferences. }
\label{fig6}
\end{figure}

\subsection{Impact of  Local Preferences and Multiple Negatives}
To assess the effectiveness of  local preferences, 
two types of DPOs are compared: 1) GPO: the standard DPO that globally ranks candidate segmentation predictions at the image level, which poses significant challenges for human raters due to varying quality of segmentation across different sub-regions of each image; 2) LPO: the local DPO that ranks candidate predictions in a patch-wise way, offering more accurate and richer information about the segmentation quality. While global preferences usually lead to inaccurate and inexact information about the segmentation quality, locally annotating preferences allows for a focus on regions with poor segmentation.

\begin{figure}[t]
\centering
\includegraphics[width=0.99\linewidth]{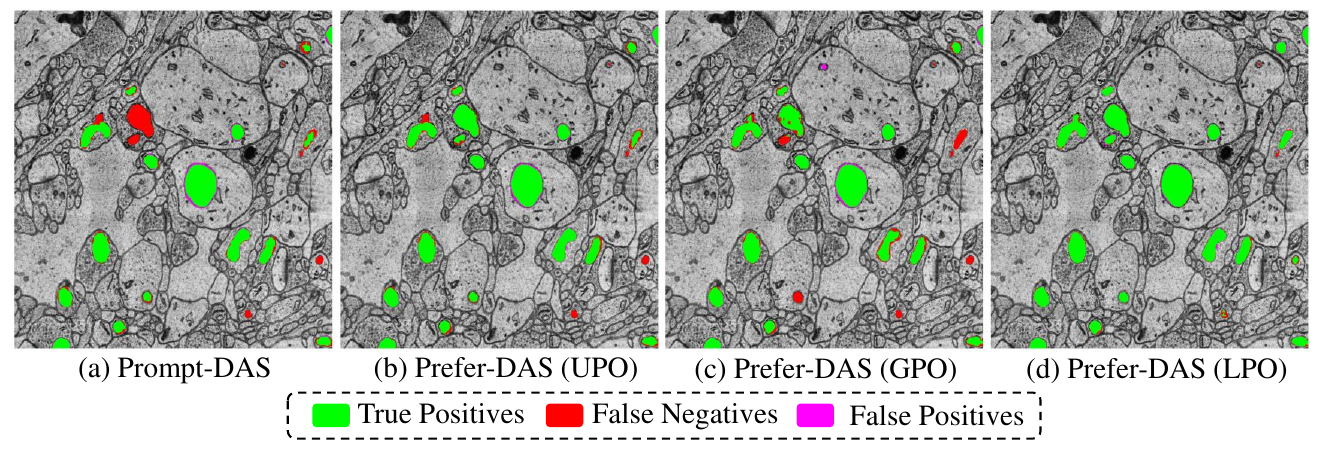}
\caption{Qualitative comparison of different strategies for preference learning. GPO refers to learning using image-level human preferences. 
} \label{fig4}
\end{figure}

As shown in Fig. \ref{fig6}, while both using global and local preferences can improve the model performance, our Prefer-DAS (LPO) model shows the best performance across all tasks. Moreover, using both human and self-learned performance can improve the model performance.
Our model using LPO outperforms the model with GPO by a large margin, when using human 
preferences.  Fig. \ref{fig4} presents a visual comparison of the Prefer-DAS with UPO, GPO, and LPO. By integrating GPO to Prompt-DAS, the Prefer-DAS (GPO) reduces some false negatives but introduces new false negatives. In contrast, the LPO  rarely introduces new false negatives while rectifying existing segmentation errors.

\begin{table}[t]
\caption{Effectiveness of using multiple negatives.}
\centering
\label{tab:3}
\setlength{\tabcolsep}{3pt}
 \renewcommand\arraystretch{1.2}
\begin{tabular}{cccccccccccccc}
\toprule
\multirow{2}{*}{DPO}&  \multicolumn{2}{c}{Negative(s)} && \multicolumn{2}{c}{H $\rightarrow$ R} & &\multicolumn{2}{c}{R $\rightarrow$ H} & & \multicolumn{2}{c}{H $\rightarrow$ Lucchi++}\\
 \cmidrule{2-3} \cmidrule{5-6} \cmidrule{8-9}\cmidrule{11-12}& Single & Multiple && Dice & PQ  & & Dice& PQ  & & Dice & PQ \\
\midrule
 \multirow{2}{*}{GPO}        & \checkmark &&&    94.0 & 78.2 & & 91.1 & 73.3 & & 93.2 & 80.5\\
 &  &\checkmark && 94.3 & 78.7 & & 91.5 & 74.2 & & 93.4 & 81.7\\
 \midrule
\multirow{2}{*}{LPO}& \checkmark &  &&   94.2 & 78.0 & & 91.3 & 74.7 & & 93.9 & 82.3 \\
& & \checkmark &&   94.6 & 79.6 & & 92.2 & 75.8 & ~ & 94.1 & 82.3 \\
\bottomrule
\end{tabular}
\end{table}

 Table \ref{tab:3} summarizes the results using multiple negatives instead of a single negative in the preference data.  As shown in Table \ref{tab:3}, the performance of using both LPO and GPO can be improved by using multiple negatives, which provides richer information about the dispreferred segmentation.  Moreover,  LPO with both single negative and multiple negatives shows improved performance over the GPO across all three tasks, validating the benefit of using local preference learning.

 \begin{figure}[t]
\centering
\includegraphics[width=1\linewidth]{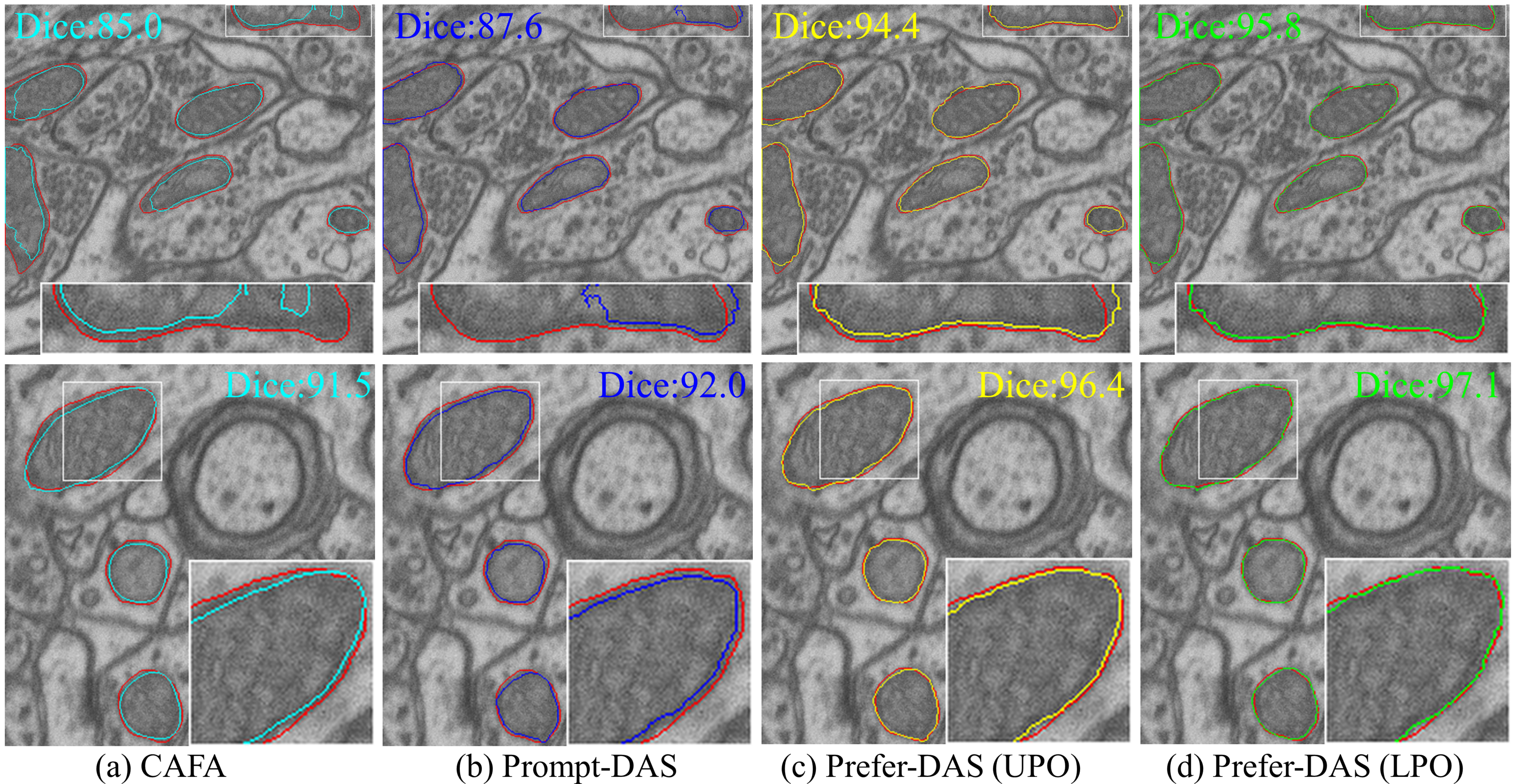}
\caption{Effectiveness of preference learning in correcting defective and biased segmentation. Red: Ground truth; Cyan: CAFA; Blue: Prompt-DAS; Yellow: Prefer-DAS (UPO); Green: Prefer-DAS (LPO). The H$\rightarrow$ Lucchi++ task is used for evaluation. Both CAFA and Prompt-DAS reveal two types of segmentation defects: 1) incomplete segmentation and 2) misalignment with target annotations, which primarily arises from the domain inconsistency in annotating inner and outer membrane edges.
}
\label{fig7}
\end{figure}
\subsection{Effectiveness of Unsupervised Preference Learning}
While the previous experiments have shown the effectiveness of supervised preference learning on reducing false positive and false negative segmentation, we further validate its effectiveness on correcting the other two types of segmentation errors: 1) incomplete mitochondria segmentation, as shown in the first row of Fig. \ref{fig7};  2) misalignment of boundaries with target annotations, as shown in Fig. \ref{fig7}. While the incomplete segmentation mainly results from the domain gap, the boundary misalignment primarily arises from annotation differences between the source and target domains.  Figure \ref{fig7} illustrates typical examples of the H$\rightarrow$Lucchi++ task.  Firstly, as shown in the first row of Fig. \ref{fig7}, the CAFA model produces incomplete segmentation of four mitochondria, whereas the Prompt-DAS has incomplete segmentation of one mitochondrion. With UPO, the Prefer-DAS (UPO) shows significantly improved segmentation, but with smaller segmentation, which is corrected by using LPO. Secondly, the boundary misalignment between the predicted segmentation and the ground truth is more obvious in the second row of Fig. \ref{fig7} due to the annotation bias on the source and target domains. Specifically, while the contours of ground truth annotations on MitoEM-Human data align with the inner membrane edges of mitochondria, the ground-truth contours of the Lucchi++ data align with the outer membrane edges of mitochondria. As shown in Fig. \ref{fig7},  both the CAFA method and our Prompt-DAS show biased segmentation similar to the source domain. Using unsupervised preference learning, our Prefer-DAS (UPO) shows significantly improved results. Moreover, learning with human preferences,  the segmentation of the Prefer-DAS (LPO) model shows strong consistency with ground truth segmentation.


\subsection{Influence of the Quantity of Inference Point Prompts}
Unlike the SAM, SAM-Med2D, Med-SAM Adapter, and WeSAM, our Prefer-DAS and Prompt-DAS can take advantage of both full and partial points as prompts. Table \ref{tab:6} summarizes the comparison results of our models on three adaptation tasks using different proportions of point prompts for inference. Overall, using both full and partial points as inference prompts can improve the segmentation performance. Using 100\% points for all instances can achieve the best results. However, assuming full points available during inference is prohibitive for processing large-scale EM stacks. As shown in Table \ref{tab:6}, using 15\% or even 0\% points, our models can achieve competitive performance on all three tasks while significantly reducing interaction requirements. Thus, our model shows wider applicability and can be applied to more scenarios with the availability of different amounts of interactions.


\section{Conclusion}
This study presents Prefer-DAS, a novel  framework that pioneers sparse promptable learning and local preference alignment for domain adaptive segmentation. Prefer-DAS allows for various configurations of utilizing point prompts and preferences, enabling it to perform  UDA and WDA and perform both automatic and interactive segmentation during inference. To effectively utilize sparse point prompts, we develop a promptable multitask model that integrates self-training and prompt-guided contrastive learning. 
 To address the dilemma between spatially varying preference alignment for semantic segmentation and image-level preference of DPO, we introduce local DPO, i.e., LPO. Additionally, we introduce UPO, which conducts unsupervised preference learning.  Comprehensive experiments conducted on four domain adaptation tasks demonstrate the effectiveness of our model across both UDA methods, WDA methods, and SAM-like methods at both automatic and interactive segmentation modes. In automatic segmentation mode, our model achieves performance that is close to or exceeds the supervised upper bound. In interactive mode, our model outperforms the supervised upper bound on three out of the four tasks. 
 
 One limitation of our method is that we have only considered one-step domain adaptation, which is challenging when new domains arrive continually.
 In future work, we will consider lightweight continuous domain adaptation with the method introduced in this study.

\begin{table}[t]
\centering
 \setlength{\tabcolsep}{1.1mm}
 \renewcommand\arraystretch{1.2}
\caption{Influence of the 
quantity of inference point prompts on interactive segmentation. Different from SAM, our models can utilize partial point prompts. }
\label{tab:6}
\begin{tabular}{cccccccccc}
\toprule
\multirow{2}{*}{Model}&Testing &\multicolumn{2}{c}{H$\rightarrow$R} & ~ & \multicolumn{2}{c}{R$\rightarrow$H} & ~ & \multicolumn{2}{c}{H$\rightarrow$Lucchi++}\\
\cmidrule{3-4}\cmidrule{6-7}\cmidrule{9-10}
 &Prompts & Dice  & PQ  & ~ & Dice & PQ & ~ & Dice  & PQ\\
 \midrule
 \multirow{4}{*}{Prompt-DAS}
&0\% & 93.3& 74.5 & ~ & 89.2 & 69.1 & & 88.2 & 73.7\\
&15\%  & 93.5& 74.2 & ~ & 90.0 & 69.8 && 88.6 & 74.7\\
&50\%  & 93.5& 74.2 & ~ & 90.4 & 70.9 && 88.6 & 74.8\\
&100\% & 93.5& 74.2 & ~ & 90.8 & 72.3 && 88.7 & 74.8\\
 \midrule
\multirow{4}{*}{}&0\% & 93.7 & 77.5 & ~ & 90.1 & 71.2 & ~ & 92.6 & 80.9\\
Prefer-DAS&15\%  & 94.1 & 78.2 & ~ & 90.6 & 70.6 & ~ & 93.2 & 82.0 \\
(UPO)&50\%  & 94.2 & 78.5 & ~ & 90.8 & 70.9 & ~ & 93.3 & 82.3 \\
&100\% & 94.3 & 79.2 & ~ & 90.9 & 71.6 & ~ & 93.4 & 82.5\\
 \midrule
\multirow{4}{*}{}&0\% & 94.6 & 79.6 & ~ & 92.2 & 75.8 & ~ & 94.1 &82.3 \\
Prefer-DAS&15\%  & 94.9 & 80.3 & ~ & 92.7 &  76.5 & ~ & 94.5 & 83.5 \\
(LPO)&50\%  & 95.0 & 80.6 & ~ & 92.8 &  76.8 & ~ & 94.6 & 84.0 \\
&100\% & 95.1 & 81.6 & ~ & 92.8 & 77.4 & ~ & 94.7 & 84.4\\
\bottomrule
\end{tabular}
\end{table}


\vfill

\end{document}